\documentclass{article}
\usepackage{spconf,amsmath,graphicx,hyperref}
\usepackage{amssymb} 
\usepackage{microtype}
\usepackage{comment}
\usepackage{booktabs,multirow}
\usepackage{stfloats}
\usepackage{hyperref}
\usepackage{fontawesome5}

\def\L{{\cal L}}

\title{Industrial Anomaly Detection via Defect-Grounded Reasoning \\ in Visual Latent Space}
\name{Jaron Yeh$^{1}$\sthanks{Equal contribution.}, Yen-Wei Chang$^{1}$\footnotemark[1], Jiang Liu$^{2}$, Shao-Yuan Lo$^{1}$}
\address{$^{1}$National Taiwan University, \, $^{2}$AMD GenAI\\
\texttt{\{b12505014, b12901068, shaoyuan\}@ntu.edu.tw, jiang.liu@amd.com}
}

\begin{document}
\ninept
\maketitle
\begin{abstract}

Industrial anomaly detection (IAD) is evolving beyond conventional detection and localization toward multimodal inspection systems that can describe, explain, and reason about fine-grained defects. Although recent multimodal large language model (MLLM)-based methods improve anomaly understanding through textual reasoning and visual guidance, they face two limitations in fine-grained inspection. First, their visual refinement often requires iteratively revisiting local image regions or augmenting with additional tools. Second, the resulting local defect evidence may not be reliably preserved throughout subsequent reasoning. To address these, we propose Anomaly-LR, a defect-grounded latent reasoning framework that first forms a global understanding of the input and then progressively refines anomaly-relevant representations directly in the visual latent space. We further construct IAD-LR-22K, the first IAD instruction dataset designed for latent reasoning, containing 22,228 image-question instances from 4,523 industrial images, with global textual reasoning traces and region-level visual annotations. Extensive experiments show that Anomaly-LR achieves state-of-the-art performance among comparable-scale methods across multiple IAD benchmarks, without requiring external references or tools. The code and data will be released at \url{https://github.com/Yen666/Anomaly-LR}.

\end{abstract}
\begin{keywords}
industrial anomaly detection, latent reasoning, multimodal large language models, instruction datasets
\end{keywords}
\section{Introduction} \label{sec:intro}

Industrial anomaly detection (IAD) is a critical component of manufacturing quality control, traditionally formulated as detecting anomalous samples and localizing defective regions. Convolutional network-based methods mainly produce anomaly scores or localization maps, offering limited semantic interpretation~\cite{li2021cutpaste,you2022unified,lo2022adversarially}. The emergence of multimodal large language models (MLLMs)~\cite{liu2023visual,hurst2024gpt,Qwen2.5-VL} has broadened industrial inspection toward more comprehensive understanding, enabling anomaly discrimination, localization, text description, and deeper analysis~\cite{anomalygpt, anomalyov2025}.


Building on this shift, recent advances introduce increasingly sophisticated reasoning mechanisms for fine-grained defect inspection. IAD-R1~\cite{li2026iadr1} enhances textual reasoning through chain-of-thought supervision and reinforcement learning. AgentIAD~\cite{agentiad2025} iteratively revisits suspicious regions through external agentic tools. Reason-IAD~\cite{reasonIAD} performs iterative latent optimization with a dynamic visual injection mechanism. Although these approaches strengthen anomaly reasoning from different perspectives, several limitations remain. Text-based reasoning provides only an indirect means of capturing fine-grained visual details, while visual refinement based on region revisiting or external visual tools introduces extra inspection steps. Moreover, the resulting local defect evidence may not always be effectively preserved and utilized throughout subsequent reasoning, as providing additional visual inputs does not necessarily guarantee that the model will attend to and exploit the relevant information~\cite{hao2024coconut,li2025lvr}. This raises the question of whether anomaly reasoning can reduce reliance on repeated visual intervention while more directly incorporating localized defect evidence into the reasoning process.

To this end, we propose Anomaly-LR, a defect-grounded latent reasoning framework that internalizes localized visual refinement within the model's latent space. Guided by textual global understanding, Anomaly-LR progressively refines intermediate hidden states by aligning them with defect-relevant visual features before answer generation. This design establishes a global-to-local reasoning process without repeated region revisiting or reliance on external tools. To train Anomaly-LR, we further construct IAD-LR-22K, the first IAD instruction dataset designed for latent reasoning, consisting of 22,228 image-question instances from 4,523 industrial images, with global textual reasoning traces and region-level visual annotations.

Extensive experiments demonstrate that Anomaly-LR consistently outperforms state-of-the-art approaches at both 3B and 7B model scales, while maintaining these gains under cross-dataset evaluation. In addition, attention map analysis shows that Anomaly-LR concentrates substantially more attention within defect regions, supporting its effectiveness in defect-grounded reasoning. This is achieved without requiring auxiliary reference images, information, or tools. We have made the Anomaly-LR model and IAD-LR-22K dataset publicly available to support future research in IAD.


Our main contributions are summarized as follows.
(1) We propose Anomaly-LR, a defect-grounded latent reasoning framework that internalizes localized visual refinement within the model's latent space.
(2) We construct IAD-LR-22K, the first IAD instruction dataset designed for latent reasoning, with global textual reasoning traces and region-level visual annotations.
(3) Experiments and attention analysis validate Anomaly-LR's effectiveness in defect-grounded reasoning without external reference images, information, or tools.

\begin{figure*}[!t]
  \centering
  \includegraphics[width=0.96\textwidth]{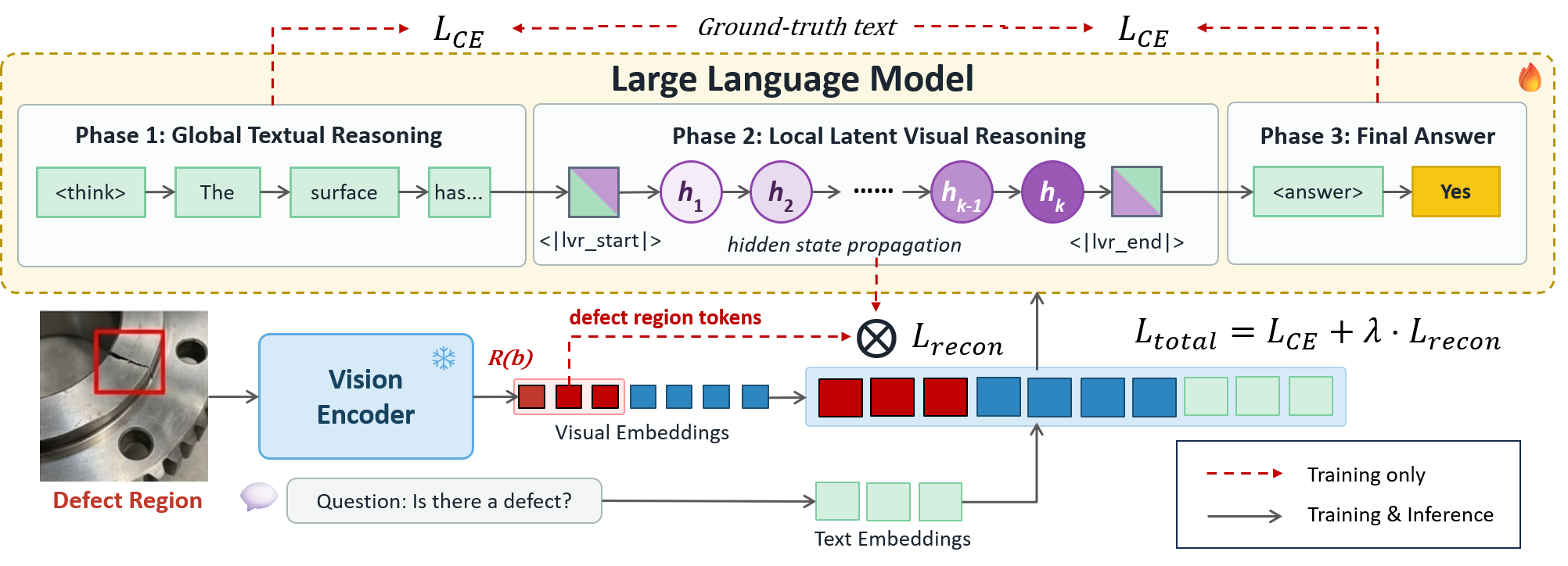}
  \vspace{-0.15 in}
  \caption{Overview of the proposed Anomaly-LR built on an MLLM backbone. Its defect-grounded reasoning process consists of three phases: global textual reasoning, $N$ steps of local latent visual reasoning delimited by the \texttt{<lvr\_start>} and \texttt{<lvr\_end>} tokens, and final answer generation. At training time, the latent positions are supervised to reconstruct the visual embeddings of the annotated defect region. At inference time, the model feeds its own hidden states back for $K$ steps.}
  \label{fig:method}
  \vspace{-0.1 in}
\end{figure*}

\section{Method} \label{sec:method}

This section introduces defect-grounded latent supervision. We supervise the intermediate states of an MLLM using visual tokens from annotated defect regions and global textual reasoning traces.

\subsection{Overall architecture} \label{sec:2.1}

Given an input image, the frozen vision encoder and projector encode it into $M$ visual embeddings $e_1,\dots,e_M$. Given a defect bounding box $b$, we define $\mathcal{R}(b) \subseteq {1,\dots,M}$ as the subset of visual tokens whose corresponding patches fall within the box, and let $N = |\mathcal{R}(b)|$.
As illustrated in Fig.~\ref{fig:method}, the proposed defect-grounded reasoning process in the LLM consists of three phases; they are global textual reasoning, local latent visual reasoning, and final answer generation:
\begin{equation}
\small
\underbrace{\texttt{<think>}t\texttt{</think>}}_{\text{global}}
\underbrace{\texttt{<lvr>}^{N}}_{\text{latent}}
\underbrace{\texttt{<answer>}ans\texttt{</answer>}}_{\text{answer}},
\label{eq:format}
\end{equation}
where $t$ is a concise global textual reasoning trace for the entire image, $ans$ is the final answer, and $\texttt{<lvr>}^{N}$ denotes $N$ latent positions delimited by the \texttt{<|lvr\_start|>} and \texttt{<|lvr\_end|>} tokens.



The global textual reasoning phase first forms a concise natural-language understanding of the entire image, after which the local latent visual reasoning phase examines suspicious regions. This global-to-local reasoning follows the natural inspection process of first identifying the object and its overall condition, and then investigating potentially defective areas~\cite{anomalyov2025}.

Inspired by~\cite{li2025lvr}, the local latent visual reasoning phase uses positions without explicit token identities and is directly supervised with visual embeddings from the annotated defect region. This encourages the model to reconstruct localized visual evidence in the latent space. The number of latent positions is determined by the size of the defect region, allowing larger defects to receive more positions for reconstruction. The final phase derives the answer from the preceding global textual and local latent visual reasoning states.

\subsection{Training and inference} \label{sec:2.2}

During training, the latent states are supervised through a latent reconstruction objective~\cite{li2025lvr}. Specifically, the input embedding $\mathbf{x}$ at the $p$-th latent position is replaced with its corresponding target embedding:
\begin{equation}
\mathbf{x}_{\pi(p)} \leftarrow e_{r_p}, \qquad r_p \in \mathcal{R}(b),
\label{eq:tf}
\end{equation}
where $\pi(p)$ is the sequence index of that position. Consequently, the model is trained to reconstruct each region token from the hidden state immediately preceding its position. Letting $h_p \triangleq h_{\pi(p)-1}$ and
$\hat{e}_p \triangleq e_{r_p}$, we employ a cosine-similarity-based reconstruction loss,
\begin{equation}
\L_{\text{recon}} \;=\; 1 - \frac{1}{N}\sum_{p=1}^{N}
\frac{\langle h_p,\, \hat{e}_p\rangle}{\lVert h_p\rVert\,\lVert \hat{e}_p\rVert}.
\label{eq:lvr}
\end{equation}
This objective differs from the $L_2$ loss used in \cite{li2025lvr}, as we observe that $h_p$ and $\hat{e}_p$ exhibit substantially different magnitudes. Hence, the squared-error objective is dominated by this scale mismatch rather than the directional alignment of the reconstructed region representation (see Sec.~\ref{sec:4.3}).

The latent reconstruction objective is combined with the standard token-level cross-entropy over the textual
positions $\mathcal{T}$, including the global reasoning trace and the final answer:
\begin{equation}
\L_{\text{CE}} \;=\; -\frac{1}{|\mathcal{T}|}\sum_{t \in \mathcal{T}}
\log p_{\theta}\!\left(y_t \mid y_{<t}\right),
\label{eq:ce}
\end{equation}
yielding the overall training objective:
\begin{equation}
\L_{\text{total}} \;=\; \L_{\text{CE}} + \lambda\,\L_{\text{recon}},
\label{eq:total}
\end{equation}
where $\lambda$ balances the two objectives. Latent positions carry no token identity and thus are excluded from $\mathcal{T}$, while $\L_{\text{recon}}$ is computed only over the latent positions. This separation also motivates the global reasoning phase. Since the answer span is short, its cross-entropy loss saturates early in training. Therefore, without the preceding global trace, $\L_{\text{recon}}$ can dominate the overall optimization, degrading answer prediction (see Sec.~\ref{sec:4.3}). Introducing a global textual reasoning trace maintains a language-modeling signal throughout training, allowing the model to learn global image understanding while preventing the latent objective from overwhelming the textual objective.

During inference, no annotations are available, so the latent segment is generated autoregressively. After emitting the \texttt{<|lvr\_start|>} token, the model feeds its own hidden state back as the next input embedding for $K$ steps before resuming textual decoding (see Fig.~\ref{fig:method}). We employ the fixed-budget decoding strategy, which was reported to outperform both a learned stopping token and an auxiliary mode-switching objective~\cite{li2025lvr}.

\subsection{IAD-LR-22K dataset} \label{sec:3.3}

Existing IAD instruction datasets primarily supervise textual predictions and reasoning processes~\cite{anomalyov2025,li2026iadr1,Jiang2024MMAD}. To our knowledge, none provides annotations that explicitly ground intermediate latent representations in localized defect evidence. To enable defect-grounded latent supervision, we construct IAD-LR-22K, the first IAD instruction dataset designed for latent reasoning. It contains 22,228 image-question instances derived from 4,523 industrial images sourced from the MMAD~\cite{Jiang2024MMAD} and Real-IAD~\cite{realiad} datasets. For MMAD, we sample 1,600 images spanning all 38 categories, with a balanced split between normal and anomalous cases. We retain the original questions, answer choices, and ground-truth answers, yielding 7,613 image-question instances. For Real-IAD, we use 2,923 images across 30 categories and generate five MMAD-style inspection questions per image, resulting in 14,615 instances. For each instance, we construct a concise global textual reasoning trace together with a region-level visual annotation. Defect regions are derived from the existing anomaly labels, while normal images are assigned object-centric regions. We refer to the resulting MMAD and Real-IAD subsets as IAD-LR-22K-MMAD and IAD-LR-22K-RealIAD, respectively. IAD-LR-22K supports defect-grounded reasoning across a range of IAD tasks, including anomaly detection, localization, defect description, and anomaly analysis.


\begin{table*}[!t]
\centering
\caption{Per-subtask accuracy on the MMAD dataset~\cite{Jiang2024MMAD}. $\ddagger$ denotes methods that adopt the same protocol of the 20\%/80\% MMAD train/test split as ours. All baseline numbers are taken from their original papers. Several methods use reference retrieval and external tool use, whereas Anomaly-LR requires neither. “Aux.”: use auxiliary reference images, domain information, or external tools; “Disc.”: Discrimination; “Cls.”: classification; “Loc.”: localization; “Desc.”: description; “Anal.”: analysis. \textbf{Bold} denotes the best value, and \underline{underline} denotes the second-best value in each column, excluding the human reference.}
\label{tab:main}
\small
\setlength{\tabcolsep}{5pt}
\begin{tabular}{ll c c c cccc cc c}
\toprule
& & & & Anomaly & \multicolumn{4}{c}{Defect} & \multicolumn{2}{c}{Object} & \\
\cmidrule(lr){5-5}\cmidrule(lr){6-9}\cmidrule(lr){10-11}
Type & Method & Scale & Aux. & Disc. & Cls. & Loc. & Desc. & Anal. & Cls. & Anal. & Average \\
\midrule
Human & Human (expert) & -- & --  & 95.04 & 75.00 & 92.31 & 83.33 & 94.20 & 86.11 & 80.37 & 86.65 \\
reference & Human (ordinary) & -- & --  & 86.90 & 66.25 & 85.58 & 71.25 & 81.52 & 89.58 & 69.72 & 78.69 \\
\midrule
Generic & GPT-4o & -- & --  & 68.63 & 65.80 & 55.62 & 73.21 & 83.41 & 94.98 & 82.80 & 74.92 \\
MLLMs & Qwen2.5-VL & 72B & --  & 72.66 & 62.31 & 67.16 & 73.56 & 81.95 & 94.30 & 86.78 & 76.96 \\
\midrule
& AnomalyGPT (AAAI'24) \cite{anomalygpt} & 7B & $\times$ & 65.57 & 27.49 & 27.97 & 36.86 & 32.11 & 29.84 & 35.82 & 36.52 \\
& AnomalyR1 (arXiv'25) \cite{AnomalyR1} & 3B & $\times$  & 60.20 & 63.50 & 70.14 & 80.47 & 85.28 & 92.48 & 86.15 & 76.96 \\
& OmniAD$^{\ddagger}$ (arXiv'25) \cite{omniad} & 3B & $\times$  & 66.30 & 75.90 & 72.90 & 65.10 & 85.40 & 93.50 & 85.60 & 77.50 \\
& OmniAD$^{\ddagger}$ (arXiv'25) \cite{omniad} & 7B & $\times$  & 68.80 & 78.80 & 75.50 & 67.20 & 86.40 & 96.00 & 86.40 & 79.87 \\
MLLM-based & AD-FM$^{\ddagger}$ (AAAI'26) \cite{adfm} & 7B & $\times$  & -- & 73.36 & 77.53 & 86.80 & 86.75 & 89.98 & 86.67 & -- \\
IAD methods & EMIT (arXiv'25) \cite{emit} & 8B & \checkmark & 73.87 & 80.85 & 76.39 & 83.00 & 85.92 & 90.26 & 83.37 & 81.95 \\
& AD-Copilot (arXiv'26) \cite{adcopilot2026} & 7B & \checkmark & \underline{73.95} & 74.29 & 76.40 & 84.92 & 86.93 & 91.86 & 87.67 & 82.29 \\
& ReasonIAD$^{*}$ (arXiv'26) \cite{reasonIAD} & 7B & \checkmark  & 70.74 & 67.97 & 60.82 & 72.78 & 82.62 & 97.17 & 84.99 & 76.73 \\
& InspectorGPT$^{\ddagger}$ (arXiv'26) \cite{InspectorGPT} & 7B & \checkmark  & 73.90 & 75.32 & 75.91 & 82.18 & \underline{88.25} & 92.94 & 88.18 & 82.38 \\
& AgentIAD$^{\ddagger}$ (arXiv'25) \cite{agentiad2025} & 3B & \checkmark  & 69.49 & 72.73 & \underline{80.94} & 85.27 & 87.84 & 93.27 & 90.59 & 82.88 \\
\midrule
Ours & Anomaly-LR$^{\ddagger}$ & 3B & $\times$ & 71.66 & \underline{82.49} & 80.57 & \underline{88.59} & 88.20 & \underline{97.43} & \underline{93.09} & \underline{86.00} \\
& Anomaly-LR$^{\ddagger}$ & 7B & $\times$  & \textbf{75.39} & \textbf{82.70} & \textbf{81.62} & \textbf{89.29} & \textbf{89.39} & \textbf{98.07} & \textbf{93.36} & \textbf{87.12} \\
\bottomrule
\end{tabular}
\end{table*}

\section{Experiments} \label{sec:expt}

\subsection{Experimental setup} \label{sec:4.1}

\noindent\textbf{Datasets and evaluation protocol.}
Our experiments cover both in-domain and out-of-domain (OOD) evaluations. For in-domain evaluation, we train Anomaly-LR on IAD-LR-22K-MMAD and test it on the remaining MMAD~\cite{Jiang2024MMAD} data, comprising 32,059 multiple-choice questions from 6,766 images. The training and test sets are disjoint, with an approximate 20\%/80\% split. This setting follows the evaluation protocol adopted in recent advances~\cite {agentiad2025,omniad,adfm}. MMAD assesses seven subtasks via multiple-choice questions: Anomaly Discrimination, Defect Classification, Defect Localization, Defect Description, Defect Analysis, Object
Classification, and Object Analysis. Accuracy is used as the metric. For OOD evaluation, we train Anomaly-LR on IAD-LR-22K-RealIAD and test it on six OOD benchmarks, including MVTec~\cite{bergmann2019mvtec}, MPDD~\cite{jezek2021deep}, VisA~\cite{zou2022spot}, DAGM~\cite{wieler2007weakly}, DTD~\cite{aota2023zero}, and SDD~\cite{tabernik2020segmentation}.


\noindent\textbf{Baselines.}
We compare our Anomaly-LR with human reference, general-purpose MLLMs~\cite{hurst2024gpt,Qwen2.5-VL}, and nine state-of-the-art MLLM-based IAD methods~\cite{anomalygpt,agentiad2025,reasonIAD,AnomalyR1,omniad,emit,adfm,adcopilot2026,InspectorGPT}.

\noindent\textbf{Implementation details.}
We adopt Qwen2.5-VL~\cite{Qwen2.5-VL} 3B and 7B as the MLLM backbones for Anomaly-LR. We train both models for three epochs with random seed 42, a batch size of 8, weight decay of 0.1, and $\lambda=0.1$, using a cosine learning-rate schedule with 3\% warmup. The learning rate is set to $1\times10^{-5}$ for the 3B model and $5\times10^{-6}$ for the 7B model. We use at most 5,120 visual tokens per image during training. At inference time, we follow the official MMAD configuration, using at most 1,280 visual tokens per image. We decode greedily with a maximum of 320 new tokens and set the number of latent reasoning steps to $K=8$.

\subsection{Main results} \label{sec:4.2}

\noindent\textbf{In-domain evaluation.}
Table~\ref{tab:main} reports per-subtask accuracy on MMAD. Anomaly-LR with both 3B and 7B backbones achieves the highest average accuracy, outperforming state-of-the-art methods. Notably, our 3B model surpasses all 7B and 8B competitors, as well as substantially larger general-purpose MLLMs, while our 7B model achieves the best accuracy across all seven subtasks. Several strong baselines, including AgentIAD~\cite{agentiad2025}, AD-FM~\cite{adfm}, and OmniAD~\cite{omniad}, adopt the same protocol of the 20\%/80\% MMAD train/test split as ours, making their results directly comparable. In addition, many methods rely on auxiliary reference images, domain information, or external tools, whereas Anomaly-LR requires neither.

\begin{table*}[!t]
\centering
\caption{OOD evaluation on six IAD benchmarks: MVTec~\cite{bergmann2019mvtec}, MPDD~\cite{jezek2021deep}, VisA~\cite{zou2022spot}, DAGM~\cite{wieler2007weakly}, DTD~\cite{aota2023zero}, and SDD~\cite{tabernik2020segmentation}. Results are reported in terms of balanced accuracy, following IAD-R1~\cite{li2026iadr1}. \textbf{Bold} denotes the best value and \underline{underline} denotes the second-best value.}
\label{tab:iadr1_main}
\setlength{\tabcolsep}{11pt}
\begin{tabular}{lcccccccc}
\toprule
\textbf{Method} & \textbf{Scale} &
\textbf{MVTec} & \textbf{MPDD} & \textbf{VisA} &
\textbf{DAGM} & \textbf{DTD} & \textbf{SDD} &
\textbf{Average} \\
\midrule

GPT-4o \cite{hurst2024gpt}          & --   & 69.6 & 60.3 & 63.5 & 63.0 & 69.9 & 65.7 & 65.3 \\
Claude Sonnet 4 \cite{Claude}  & --   & 67.6 & 65.9 & 63.5 & 69.2 & 88.4 & 81.7 & 72.7 \\


Qwen2.5-VL \cite{Qwen2.5-VL}            & 3B   & 62.6 & 52.9 & 58.4 & 54.2 & 64.4 & 50.3 & 57.1 \\
Qwen2.5-VL \cite{Qwen2.5-VL}            & 7B   & 66.0 & 56.0 & 58.4 & 57.7 & 59.2 & 67.4 & 60.8 \\
AnomalyGPT (AAAI'24) \cite{anomalygpt}  & 7B   & 46.6 & 54.2 & 57.3 & 49.6 & 64.1 & 49.5 & 53.6 \\
Anomaly-OV (CVPR'25) \cite{anomalyov2025}     & 7B   & 74.3 & \textbf{70.3} & \underline{74.3} & 77.5 & 90.7 & 88.7 & 78.9 \\
IAD-R1 (AAAI'26) \cite{li2026iadr1}     & 3B   & 77.6 & 59.2 & 69.8 & 85.2 & 89.1 & 83.4 & 77.4 \\
IAD-R1 (AAAI'26) \cite{li2026iadr1}      & 7B   & \textbf{81.9} & 65.8 & \textbf{75.4} & 85.2 & \underline{90.8} & 83.4 & \underline{80.4} \\
\midrule


Anomaly-LR
& 3B
& 74.9 & 66.6 & 63.4 & \underline{92.3} & 85.5 & \underline{91.5} & 79.0 \\
Anomaly-LR
& 7B
& \underline{77.9} & \underline{66.8} & 67.3 & \textbf{95.4} & \textbf{92.8} & \textbf{92.7} & \textbf{82.2} \\

\bottomrule
\end{tabular}

\end{table*}
\begin{table}[!t]
\centering
\vspace{-0.15 in}
\caption{Ablation results on the MMAD dataset~\cite{Jiang2024MMAD}. All models are trained using the same protocol.}
\label{tab:component}
\small
\setlength{\tabcolsep}{12pt}
\begin{tabular}{lcc}
\toprule
Configuration & 3B & 7B \\
\midrule
\multicolumn{3}{l}{\emph{(a) Architecture design}}\\
Qwen2.5-VL backbone~\cite{Qwen2.5-VL} & 65.25 & 71.07 \\
Phase 1 + Phase 3 & 85.49 & 87.03 \\
Phase 2 + Phase 3 & 83.37 & 84.01 \\
Phase 1 + Phase 2 + Phase 3 (ours) & \textbf{86.00} & \textbf{87.12} \\
\midrule
\multicolumn{3}{l}{\emph{(b) Latent reconstruction loss ($\L_{\text{recon}}$)}}\\
$L_2$-norm & 85.23 & \textbf{87.13} \\
Cosine similarity (ours) & \textbf{86.00} & 87.12 \\
\midrule
\multicolumn{3}{l}{\emph{(c) Latent budget during inference ($K$)}}\\
$K=0$ (no rollout) & 84.77 & 86.25 \\
$K=8$ (default) & \textbf{86.00} & \textbf{87.12} \\
\bottomrule
\end{tabular}
\end{table}

\noindent\textbf{OOD evaluation.}
Table~\ref{tab:iadr1_main} reports the OOD evaluation results. IAD-R1~\cite{li2026iadr1} uses the same Qwen2.5-VL backbone and Real-IAD~\cite{realiad} training images as Anomaly-LR, making it directly comparable. Anomaly-LR improves the average accuracy from 77.4\% to 79.0\% at the 3B scale and from 80.4\% to 82.2\% at the 7B scale, achieving state-of-the-art performance. These results show the cross-dataset generalizability of the proposed defect-grounded latent reasoning framework.


\subsection{Ablation study} \label{sec:4.3}

\noindent\textbf{Architecture design.}
Table~\ref{tab:component}a compares different architectural variants on MMAD~\cite{Jiang2024MMAD} under the in-domain evaluation setting. The Phase 1 + Phase 3 variant removes the local latent visual reasoning stage and is trained only with $\L_{\text{CE}}$, making it analogous to standard text-based supervised fine-tuning. The Phase 2 + Phase 3 variant retains local latent visual reasoning but removes the global textual reasoning stage, making it similar to~\cite{li2025lvr}. The full three-phase architecture achieves the best performance. In particular, Phase 2 + Phase 3 tends to collapse to a single answer because the answer span is short and its cross-entropy loss saturates early during training. These results show that global textual reasoning and local latent visual reasoning are both effective and complementary.

\noindent\textbf{Latent reconstruction loss.}
Table~\ref{tab:component}b compares different objectives for the latent reconstruction loss $\L_{\text{recon}}$. Cosine similarity outperforms the $L_2$ loss used in~\cite{li2025lvr} at the 3B scale and performs comparably at the 7B scale. We attribute this to the scale mismatch between the latent hidden states and target visual embeddings. The $L_2$ loss is sensitive to differences in magnitude, whereas cosine similarity focuses on their directional alignment.

\noindent\textbf{Latent budget during inference.}
As mentioned in Sec.~~\ref{sec:4.1}, we set $K=8$ as the default number of latent reasoning steps. As shown in Table~\ref{tab:component}c, removing the latent rollout entirely ($K=0$) drops the accuracy, confirming that iterative reasoning in the visual latent space contributes to the final prediction.


\begin{figure}[!t]
  \centering
  \includegraphics[width=0.96\columnwidth]{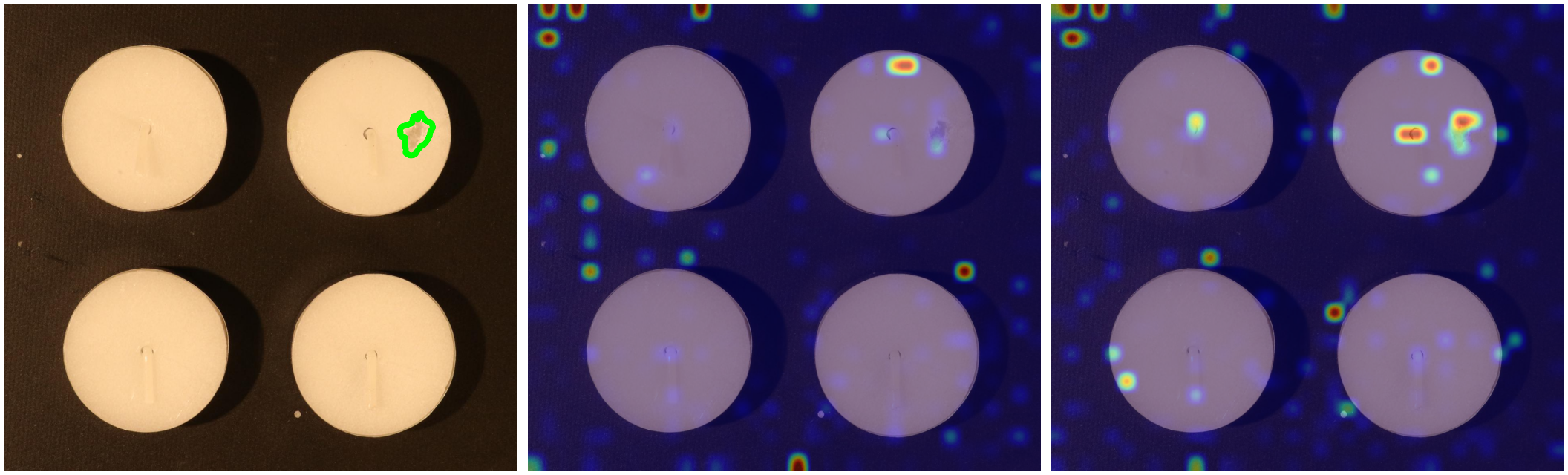}\\[1pt]
  \makebox[0.3333\columnwidth]{\small Input}%
  \makebox[0.3333\columnwidth]{\small Qwen2.5-VL-3B}%
  \makebox[0.3333\columnwidth]{\small Anomaly-LR-3B}
  \caption{Attention maps of the models on a VisA~\cite{zou2022spot} test image (candle). The green circle denotes the ground-truth defect region. Compared to the Qwen2.5-VL-3B~\cite{Qwen2.5-VL} backbone, Anomaly-LR-3B focuses much more strongly on the defect region.}
  \label{fig:attn}
\end{figure}

\noindent\textbf{Attention map analysis.}
Fig.~\ref{fig:attn} visualizes the attention maps of the models on a VisA~\cite{zou2022spot} test image (candle). Compared to the Qwen2.5-VL-3B~\cite{Qwen2.5-VL} backbone, Anomaly-LR-3B focuses much more strongly on the defect region. Following the evidence-attention diagnostic of~\cite{ease}, we quantify each map by the attention mass assigned to the annotated defect region at layer $\lfloor 2L/3 \rfloor$, normalized by the mass that uniform attention would assign to the same region. Qwen2.5-VL-3B assigns $4.4\times$ the uniform attention mass to the defect region, whereas Anomaly-LR-3B reaches $19.5\times$. These results demonstrate that the proposed defect-grounded latent reasoning more effectively directs the model's attention toward localized defect evidence.

\section{Conclusion}

We introduce Anomaly-LR, a defect-grounded latent reasoning framework that internalizes localized visual refinement within the model's latent space, and IAD-LR-22K, the first IAD dataset designed for latent reasoning. Guided by global textual reasoning, Anomaly-LR progressively refines intermediate hidden states by aligning them with defect-relevant visual features. This design reduces reliance on repeated visual intervention. Anomaly-LR achieves strong performance across model scales and benchmarks, demonstrating the effectiveness of defect-grounded latent reasoning for fine-grained industrial anomaly understanding.


\section{Compliance with Ethical Standards}
This study uses existing public benchmarks and collects no new human-subject data. No ethical approval was required.


\bibliographystyle{IEEEbib}
\bibliography{refs}

\end{document}